\documentclass[conference]{IEEEtran}
\IEEEoverridecommandlockouts
\usepackage{cite}
\usepackage{amsmath,amssymb,amsfonts}
\usepackage{graphicx}
\usepackage{textcomp}
\usepackage{xcolor}
\usepackage{subcaption}
\usepackage{amsmath}
\usepackage{amsfonts}
\usepackage{multirow}
\usepackage{multicol}
\usepackage{booktabs}
\usepackage{balance}
\usepackage{graphicx}
\usepackage{tabularx}
\usepackage[linesnumbered,ruled,vlined]{algorithm2e}
\def\BibTeX{{\rm B\kern-.05em{\sc i\kern-.025em b}\kern-.08em
    T\kern-.1667em\lower.7ex\hbox{E}\kern-.125emX}}

\makeatletter
\def\ps@IEEEtitlepagestyle{%
  \def\@oddfoot{\mycopyrightnotice}%
  \def\@evenfoot{}%
}
\def\mycopyrightnotice{%
  {\footnotesize 978-1-6654-0126-5/21/\$31.00 \textcopyright2021 IEEE\hfill}
  \gdef\mycopyrightnotice{}
}

\begin{document}

\title{Textual Data Augmentation for Patient Outcomes Prediction
}

\author{\IEEEauthorblockN{Qiuhao Lu}
\IEEEauthorblockA{\textit{University of Oregon}\\
Eugene, OR, USA \\
luqh@cs.uoregon.edu}
\and
\IEEEauthorblockN{Dejing Dou}
\IEEEauthorblockA{\textit{University of Oregon}\\
Eugene, OR, USA}
\IEEEauthorblockA{\textit{Baidu Research}\\
Beijing, China \\
dou@cs.uoregon.edu}
\and
\IEEEauthorblockN{Thien Huu Nguyen}
\IEEEauthorblockA{\textit{University of Oregon}\\
Eugene, OR, USA \\
thien@cs.uoregon.edu}
}

\maketitle

\begin{abstract}
Deep learning models have demonstrated superior performance in various healthcare applications. However, the major limitation of these deep models is usually the lack of high-quality training data due to the private and sensitive nature of this field. In this study, we propose a novel textual data augmentation method to generate artificial clinical notes in patients' Electronic Health Records (EHRs) that can be used as additional training data for patient outcomes prediction. Essentially, we fine-tune the generative language model GPT-2 to synthesize labeled text with the original training data. More specifically, We propose a teacher-student framework where we first pre-train a teacher model on the original data, and then train a student model on the GPT-augmented data under the guidance of the teacher. We evaluate our method on the most common patient outcome, i.e., the 30-day readmission rate. The experimental results show that deep models can improve their predictive performance with the augmented data, indicating the effectiveness of the proposed architecture.

\end{abstract}


\section{Introduction}
Patient outcomes, including patients' readmission risk, mortality rate, and length of stay (LOS), have been examined as important measurements for evaluating the quality of hospital care \cite{davison2016patient}. As the most commonly reported health outcome in the United States, readmissions are estimated to cost Medicare \$15 billion annually, of which \$12 billion is potentially preventable, according to the Medicare Payment Advisory Committee \cite{hackbarth2009reforming}. This highlights the importance of identifying patients at high risk of readmission.

Over the past few years, there has been a surge of interest in making predictions on patient outcomes using deep learning techniques, such as readmission prediction \cite{lin2019analysis}, mortality prediction \cite{Harutyunyan2017MultitaskLA}, length of stay prediction \cite{ma2020length}, etc. Most of these studies heavily rely on feature engineering, where they select statistically significant features from patients' Electronic Health Records (EHRs), and feed them into deep models like a LSTM-CNN network \cite{lin2019analysis}.

A common theme among these studies is that they all rely on numerical and time-series features of patients, while neglecting the clinical notes of EHRs which prove to be informative in such predictive tasks. This motivates recent studies to cast this task as text classification, where the contextual content of EHRs are leveraged to make predictions. For example, Lu \textit{et al.} propose a graph-based method that converts clinical notes to multi-view graphs and use them to predict ICU patients' 30-day unplanned readmission risk, surpassing state-of-the-art numerical-based methods \cite{lu2021predicting}.

However, in real-world downstream applications, deep learning models often suffer from data limitation as they require large amounts of data for effective training. The situation is even worse in the biomedical domain due to the private and sensitive nature of this field. Despite data shortage, data imbalance is also an issue for patient outcomes prediction, e.g., only few patients are readmitted post-discharge. These data issues make patient outcomes prediction more challenging than general predictive tasks.

A natural solution to these problems is data augmentation, where new data is synthesized based on existing training data. This strategy has been actively applied in the field of computer vision, where researchers alter the training images to create a larger dataset by introducing random transformations such as translation, mirroring, rotation, and more \cite{mclaughlin2015data}. However, these augmentation strategies that are successful in computer vision cannot be easily applied to textual data due to the inherent complexity of natural language \cite{amin-nejad-etal-2020-exploring}, where the grammatical or semantic consistency of text could hardly be preserved after transformation \cite{Anaby-Tavor_Carmeli_Goldbraich_Kantor_Kour_Shlomov_Tepper_Zwerdling_2020}. As to the specific task of readmission prediction, such issues, e.g., data imbalance, are either ignored \cite{lu2019learning} or processed with sampling techniques \cite{junqueira2019machine}, such as SMOTE \cite{chawla2002smote} or ROSE \cite{menardi2014training} that do not cope with textual data.

Recently, natural language generation (NLG) techniques have been leveraged as a new means for textual data augmentation. With the development of large pre-trained generative language models like GPT-2 \cite{radford2019language}, researchers are able to generate high-quality and semantic-consistent textual data while preserving the annotated labels. This augmentation strategy has been applied in various NLP downstream tasks, such as event detection \cite{veyseh2021unleash}, relation extraction \cite{Papanikolaou2020DAREDA}, commonsense reasoning \cite{yang2020g}, etc. However, in the biomedical field, leveraging GPT-2 to facilitate clinically-relevant predictive models is under-explored.

One main challenge of using GPT-2 for textual data augmentation is noise control. Existing studies typically address this issue in a isolated way, where they introduce heuristic filtering mechanisms to eliminate low-quality samples \cite{Anaby-Tavor_Carmeli_Goldbraich_Kantor_Kour_Shlomov_Tepper_Zwerdling_2020} and feed the rest to the downstream model. However, such filtering strategies are prone to coverage errors and thus inevitably make incorrect judgements on the generated samples \cite{veyseh2021unleash}, which would cause false inclusion of good samples or false exclusion of bad samples. Moreover, the combined data samples are treated equally by the to-be-trained downstream model, and this would negatively impact the model as a consequence.

To overcome this issue, we propose a conceptually different strategy where all the generated samples are involved during training. We preserve all the generated samples in the first place, and then introduce a teacher-student framework to regularize the representation learning of the generated samples with knowledge transferred from the original data. More specifically, we pre-train a teacher model on the original data and then train a student model on the combined data adaptively under the guidance of the teacher. The goal is to transfer the knowledge learned in the teacher model into the student model by enforcing a knowledge consistency between them, and that eventually the student model can be improved. We evaluate the framework with the state-of-the-art textual-based readmission prediction model \cite{lu2021predicting}, the results of which indicates the effectiveness of the method.

The contributions of this work can be summarized as follows:
\begin{itemize}
    \item We propose a novel architecture that leverages GPT-2 for \textbf{Med}ical text \textbf{Aug}mentation (MedAug) in the task of patient outcomes prediction. Essentially, we introduce a teacher-student framework that aims to control the noise of the generated text by enforcing a knowledge consistency across the original and artificial texts.
    \item Taking the readmission prediction task as a case study, we specifically investigate the performance of MedAug with the state-of-the-art readmission prediction model as well as a baseline model. Extensive experiments demonstrate that both models can improve their performance with the augmented data, indicating the effectiveness of the proposed architecture.
\end{itemize}

\section{Methodology}
\label{methodology}
\subsection{Notations}
In this study, we focus on textual-based readmission prediction models where the prediction task is cast as a supervised binary text classification problem. We refer to the original training dataset as $D_{train} = \{(x_1,y_1), (x_2,y_2), \dots, (x_n,y_n)\}$ where $x_i$ is a clinical note and $y_i\in\{0,1\}$ indicates whether the patient is readmitted or not. Note that $D_{train}$ is imbalanced where negative samples are 3x more than the positive ones, as only few patients are readmitted post-discharge. We similarly denote the test set by $D_{test}$ and the validation set by $D_{valid}$. We also denote the synthesized training set by $D_{synthetic}$, which is generated by the fine-tuned GPT-2 model $\mathcal{G}_{tuned}$. We also combine the original and generated training data together to create a large training dataset $D_{combined}=D_{train}\cup D_{synthetic}$. Finally, we refer to the prediction method as $\mathcal{M}$.

\subsection{Data Generation}
We fine-tune the GPT-2 model $\mathcal{G}$ on the original training data $D_{train}$ so that it can synthesize reasonable textual data that can be used for the training of $\mathcal{M}$. To preserve the class information, we prepend the class label $y_i$ to each note in the training data, i.e., $y_i\texttt{SEP}x_i\texttt{EOS}$, where \texttt{SEP} and \texttt{EOS} are the separation and ending token, respectively. We then fine-tune GPT-2 on the processed training data with the objective of predicting the next token, the same way it was pre-trained \cite{radford2019language}. The fine-tuned model is regarded as $\mathcal{G}_{tuned}$.

For generating new data, we use the class label along with a short context as the prompt to $G_{tuned}$, i.e., $\texttt{prompt} = y_1\texttt{SEP}w_1w_2$ where the first two tokens are included as context, as suggested in \cite{Anaby-Tavor_Carmeli_Goldbraich_Kantor_Kour_Shlomov_Tepper_Zwerdling_2020}. Since in our case the negative samples are 3x more than the positive ones, we only focus on generating positive samples to fulfill the gap, i.e., only the positive label $y_1$ is used for generation. We denote the generated training data by $D_{synthetic}$.

\subsection{Data Integration}
As mentioned in the introduction, noise control is one of the main challenges for textual data augmentation. In this work, we propose a teacher-student framework for data integration so that all the generated samples are included for training. We first pre-train a teacher prediction model $\mathcal{M}_{teacher}$ on $D_{train}$ to capture the inherent knowledge of the original clean training data. Then we train the student model $\mathcal{M}_{student}$ on the combined data $D_{combined}$ in a way that the teacher's knowledge can be used to guide the student learning. To achieve this, we aim to enforce a knowledge consistency between the student and the teacher, by incorporating a KL divergence penalty to push the representations learned in the student model close to that in the teacher. Essentially, we seek to jointly minimize the KL divergence between the predicted label probability distribution of the student and the teacher, along with the original training objective of the student, i.e., $\mathcal{L} = \mathcal{L}_{student} + \tau \mathcal{L}_{KL}$. It's also worth mentioning that in this study we use the KL divergence to control noise in the labeled data generated by GPT-2, which is different from knowledge distillation on unlabeled data \cite{44873}. The architecture is defined in Algorithm~\ref{medaug}.


\begin{algorithm}[bt!]
\KwIn{$D_{train}$, $\mathcal{G}$, $\mathcal{M}$}
\KwOut{$\mathcal{M}_{student}$}
Fine-tune $\mathcal{G}$ on $D_{train}$ to obtain $\mathcal{G}_{tuned}$\\
Use $\mathcal{G}_{tuned}$ to generate $D_{synthetic}$ and combine it with $D_{train}$ to obtain $D_{combined}$\\
Pre-train a teacher model $\mathcal{M}_{teacher}$ on $D_{train}$\\
Train the student model $\mathcal{M}_{student}$ on $D_{combined}$ under the guidance of $\mathcal{M}_{teacher}$\\
\textbf{Return} $\mathcal{M}_{student}$
\caption{MedAug}\label{medaug}
\end{algorithm}

\section{Experiments}
In this section, we evaluate the proposed framework on the task of ICU patients readmission prediction where we aim to show the effectiveness of MedAug. Essentially, we take as input the clinical note of patients' EHRs, and predict whether or not the patient will be readmitted within 30 days after discharge or transfer.

\subsection{Dataset}
The experiment is conducted based on the MIMIC-III Critical Care (Medical Information Mart for Intensive Care III) Database, which is a large, freely-available database composed of de-identified EHR data \cite{johnson2016mimic}. Following prior work \cite{zhang2020learning}, we extract the \texttt{Discharge Summaries} from EHRs as the data. For a fair comparison, we use the same data split with the baseline \cite{lu2021predicting} where $48,393$ generated documents are split into training ($80\%$), validation ($10\%$), and testing ($10\%$). Specifically, the original training set $D_{train}$ consists of $7555$ positive samples and $30247$ negative samples which are denoted by $D_{train,1}$ and $D_{train,0}$, respectively.

\subsection{Evaluation Metrics}
We follow the prior work \cite{lu2021predicting} and use the area under the receiver operating characteristics curve (AUROC), the area under the precision recall curve (AUPRC), and the recall at precision of $80\%$ (RP80) for evaluation.

\subsection{Prediction Models}
We consider the following two prediction models for evaluation in this experiment. We evaluate with two prediction models to investigate how MedAug performs when equipped with a base model and an advanced model.
\begin{itemize}
    \item ClinicalBERT. ClinicalBERT is a domain-specific BERT variant initialized from BioBERT v1.0 \cite{lee2020biobert} and pre-trained on MIMIC notes \cite{alsentzer-etal-2019-publicly}. In this study, we add a linear classification head on top of it and use it as a baseline.
    \item MedText. MedText is a textual-based readmission prediction model and reports state-of-the-art performance on this task \cite{lu2021predicting}.
\end{itemize}

\subsection{Augmentation Baselines}
We consider two augmentation baselines for comparison.
\begin{itemize}
    \item base. The base strategy is a baseline that all generated samples are included while no noise control is applied.
    \item LAMBADA. LAMBADA is an augmentation method specified for text classification \cite{Anaby-Tavor_Carmeli_Goldbraich_Kantor_Kour_Shlomov_Tepper_Zwerdling_2020}. Basically, they pre-train a classifier on the clean data and use it to select confident samples.
\end{itemize}

\subsection{Results}
Table~\ref{table1} shows the test performance of the two readmission prediction models, along with three augmentation strategies. We observe that without controlling the noise, i.e., base, both models demonstrate inferior performance, indicating the non-negligible level of noise in the generated samples. On the other hand, with MedAug, both models demonstrate better performance, and the improvement is significant comparing with the other two baselines, indicating the effectiveness of this framework.

\begin{table}[t]
\begin{center}
\caption{Test performance on 30-day unplanned ICU patient readmission prediction.}
\label{table1}
\begin{tabular}{lccc}
    \toprule
    Method  & AUROC & AUPRC & RP80 \\
    \midrule
    ClinicalBERT   &	0.782&	0.549 & 0.201\\
    ClinicalBERT-base  &	0.779&	0.550 & 0.221\\
    ClinicalBERT-LAMBADA &	0.782&	0.543 & 0.196\\
    ClinicalBERT-MedAug  &	\textbf{0.791}&	\textbf{0.565} & \textbf{0.234}\\
    \midrule
    MedText   &\textbf{0.823}&	0.632 & 0.319\\
    MedText-base  &	0.803&	0.599 & 0.290\\
    MedText-LAMBADA   &0.806&	0.604 & 0.266\\
    MedText-MedAug  &	0.822&	\textbf{0.633} & \textbf{0.328}\\    
    \bottomrule
\end{tabular}
\end{center}
\end{table}

\begin{table}[t]
\begin{center}
\caption{Influence of $|D_{synthetic}|$ by MedAug.}
\label{table2}
\begin{tabular}{lcccc}
    \toprule
    $|D_{synthetic}|$ & Method  & AUROC & AUPRC & RP80 \\
    \midrule
    3k & ClinicalBERT   &	0.777&	0.550 & 0.220\\
    9k & ClinicalBERT  &	0.784&	0.567 & 0.246\\
    12k & ClinicalBERT &	0.784&	0.569 & 0.245\\
    24k & ClinicalBERT  &	0.783&	0.566 & 0.251\\
    \midrule
    3k & MedText   &0.812&	0.621 & 0.329\\
    9k & MedText   &0.811&	0.623 & 0.337\\
    12k & MedText  &0.806&	0.611 & 0.311\\
    24k & MedText  &0.809&	0.618 & 0.331\\    
    \bottomrule
\end{tabular}
\end{center}
\end{table}


\section{Analysis}
In this section, we investigate three potential issues that might have influenced the performance of MedAug, i.e., the number of synthesized samples $|D_{synthetic}|$, the fine-tuning and generation strategy for GPT-2, and the version of GPT-2.

\subsection{Number of Synthesized Samples}

Table~\ref{table2} shows the validation performance of different $|D_{synthetic}|$, demonstrating the influence of the size of the synthetic training set. With the increasing of synthesized samples, the general performance appears to have reached a peak and then begin to drop slightly. We conjecture that there is a trade-off between the size and the performance, and it is determined by the augmentation strategy.

\subsection{GPT-2 Fine-tuning Strategy}
It is common that patient outcomes demonstrate an imbalanced distribution, e.g., only few patients are readmitted after discharge. In our case, negative samples are 3x more than the positive ones, i.e., $D_{train,0}=4\times D_{train,1}$. Therefore, when fine-tuning GPT-2 using the original training data, we explicitly make it balanced to prevent the negative samples from misleading GPT-2, by performing random under-sampling over $D_{train}$. As to the prompt to GPT-2 in generating new samples, we compare two options, i.e., w/ and w/o context, where context refers to the first two tokens of the text.

We investigate the two issues and show the comparison results on the validation set in Table~\ref{table3}. Note that to avoid the impact from augmentation strategies, we use the base method, i.e., simply include all the samples, in this experiment. Generally, a balanced training set and a prompt with context are the best options for fine-tuning and generation with GPT-2 in this task.

\subsection{GPT-2 Version}
Finally, we investigate the version of GPT-2 and its influence over the quality of synthesized samples. We test with GPT-2-small and GPT-2-medium and show the results in Table~\ref{table4}. Generally, we observe that GPT-2-medium has a minor advantage over GPT-2-small. However, considering the training cost and efficiency, we choose to use GPT-2-small for all the experiments in this study.

\begin{table}[t]
\begin{center}
\caption{Influence of GPT-2 fine-tuning/generation strategies.}
\label{table3}
\begin{tabular}{lccccc}
    \toprule
    Prompt & Balanced & Method  & AUROC & AUPRC & RP80 \\
    \midrule
    w/o ctx &N & ClinicalBERT   &	0.771&	0.535 & 0.205\\
    w/o ctx &Y & ClinicalBERT  &	0.773&	0.536 & 0.216\\
    w/ ctx &N & ClinicalBERT &	0.767&	0.531 & 0.198\\
    w/ ctx &Y & ClinicalBERT  &	0.775&	0.551 & 0.226\\
    \midrule
    w/o ctx &N & MedText   &0.791&	0.589 & 0.296\\
    w/o ctx &Y & MedText   &0.791&	0.595 & 0.313\\
    w/ ctx &N & MedText  &0.791&	0.593 & 0.296\\
    w/ ctx &Y & MedText  &0.795&	0.602 & 0.318\\    
    \bottomrule
\end{tabular}
\end{center}
\end{table}

\begin{table}[t]
\begin{center}
\caption{Influence of the version of GPT-2.}
\label{table4}
\begin{tabular}{lcccc}
    \toprule
    GPT-2 version & Method  & AUROC & AUPRC & RP80 \\
    \midrule
    small & ClinicalBERT   &	0.784&	0.567 & 0.246\\
    medium & ClinicalBERT  &	0.783&	0.568 & 0.252\\
    \midrule
    small & MedText &	0.811&	0.623 & 0.337\\
    medium & MedText  &	0.811&	0.623 & 0.339\\
    \bottomrule
\end{tabular}
\end{center}
\end{table}

\section{Related Work}
Readmission prediction is challenging task and has attracted a lot of attention over the years. Lin \textit{et al.} select numerical chart event features over a 48-hour time window and feed them to a deep LSTM-CNN network \cite{lin2019analysis} and achieve much better performance than traditional methods. Zhang \textit{et al.} propose CC-LSTM that encodes external knowledge into text representations and outperforms Lin's work \cite{zhang2020learning}. Afterwards Lu \textit{et al.} propose to convert clinical notes to multi-view graphs and process them with graph convolution networks \cite{lu2021predicting}. These studies demonstrate the value of textual content in EHRs and motivate us to apply textual data augmentation to this task.

Recently, using GPT-2 for augmenting textual training data has been studied for a variety of tasks in the NLP field, such as such as event detection \cite{veyseh2021unleash}, relation extraction \cite{Papanikolaou2020DAREDA}, commonsense reasoning \cite{yang2020g}, spoken language understanding \cite{peng2020data}, extreme multi-label classification \cite{zhang2020data}, etc. However, none of these works has leveraged GPT-2 for patient outcomes prediction. This highlights the importance of this study and motivates us to explore more of this direction.

\section{Conclusion}
In this paper, we propose MedAug, a framework that leverages GPT-2 to synthesize artificial training data for patient outcomes prediction. We evaluate the method on task of ICU patients readmission prediction, the results of which demonstrate that either a baseline or an advanced prediction model can benefit from the synthesized training data, under the framework of MedAug. Essentially, to control the noise in the synthesized data, we propose a teacher-student architecture that enforces a knowledge consistency across the original and artificial texts. We introduce a mechanism for knowledge consistency enforcement to mitigate noises from generated data based on KL divergence.

On the other hand, as a preliminary exploration of this direction, we do observe that the improvement for the advanced model is less significant than the baseline model, which motivates us to investigate further in the future work.

\section*{Acknowledgment}
This research has been supported by the Army Research Office (ARO) grant W911NF-21-1-0112 and the NSF grant CNS-1747798 to the IUCRC Center for Big Learning. We also would like to thank the IBM-Almaden research group for their support in this work. This research is also based upon work supported by the Office of the Director of National Intelligence (ODNI), Intelligence Advanced Research Projects Activity (IARPA), via IARPA Contract No. 2019-19051600006 under the Better Extraction from Text Towards Enhanced Retrieval (BETTER) Program. The views and conclusions contained herein are those of the authors and should not be interpreted as necessarily representing the official policies, either expressed or implied, of ARO, ODNI, IARPA, the Department of Defense, or the U.S. Government. The U.S. Government is authorized to reproduce and distribute reprints for governmental purposes notwithstanding any copyright annotation therein. This document does not contain technology or technical data controlled under either the U.S. International Traffic in Arms Regulations or the U.S. Export Administration Regulations.

\bibliography{cites.bib}
\balance
\bibliographystyle{IEEEtran}

\end{document}